\documentclass{article}
\usepackage{spconf,amsmath,graphicx, amssymb, amsfonts}
\usepackage{algorithm}
\usepackage{algorithmic}
\usepackage{subfigure}
\usepackage{booktabs}

\title{On Network Science and Mutual Information for Explaining Deep Neural Networks}
\name{Brian Davis$^{1}\sthanks{Equal Contribution}$, Umang Bhatt$^{1,2*}$, Kartikeya Bhardwaj$^{1,3*}$, Radu Marculescu$^{1,4}$, and Jos\'e M.F. Moura$^{1}$}
\address{ $^{1} $Carnegie Mellon University,  
$^{2} $University of Cambridge, 
$^{3} $Arm Inc.,
$^{4} $The University of Texas at Austin}
\begin{document}
%
\maketitle

\begin{abstract}
In this paper, we present a new approach to interpret deep learning models. By coupling mutual information with network science, we explore how information flows through feedforward networks. We show that efficiently approximating mutual information allows us to create an information measure that quantifies how much information flows between any two neurons of a deep learning model. To that end, we propose NIF, Neural Information Flow, a technique for codifying information flow that exposes deep learning model internals and provides feature attributions.
\end{abstract}
\section{Introduction}
\label{sec:intro}
As deep learning gains popularity, there has been an influx in methods that attempt to explain how deep learning begets its predictive power. Most approaches to interpret deep neural networks are model agnostic and make a local approximation in the feature space around the datapoint to be explained \cite{lime}. However, such techniques fail to capture global \textit{model-specific} behavior that is crucial to understand if the function learned by a deep learning model aligns well with human intuition. Moreover, current noisy approximations neglect the topological structure of the model used for prediction \cite{int}. 

It is easy to forget the network structure of deep learning models, particularly feed forward models, that resemble directed acyclic graphs. However, understanding the topological structure of different models can not only help decide the architecture best suited for the task at hand but also help expose the internal interactions between neurons at inference time. While the existing interpretability techniques \cite{l2x} shed light on which input features are responsible for a given prediction, prior art still fails to quantify how information flows through a deep network at the neuron-level. This prevents answering one of the most fundamental questions in deep learning: \textit{How much information flows through a deep network from input features to each subsequent neuron?} 

To address this question, we consider two types of interpretability notions: (\textit{i}) model interpretability via attribution to input features, and (\textit{ii}) network architecture interpretability with respect to how information flows from neuron to neuron for a given pretrained model. We believe, addressing notion (\textit{ii}) from a fundamental information theory standpoint will automatically reveal insights about the precise decision-making process followed by the model (i.e., notion (\textit{i})). 

Using an information theoretic measure, we model the flow of information via \textit{Neural Information Flow} (NIF) between neurons in consecutive layers to expose how simple deep learning models can learn complex functions of input features.
We further analyze information flow between neurons from a network science~\cite{networksci} perspective, where each neuron in the deep network becomes a node in the network. 
NIF recovers an information-theoretic feature attribution, namely a rank of feature importance to a given class. 
Combining an information measure with the ability to propagate information through the network can help us visualize information flow and feature attributions simultaneously. 

\section{Background}
\subsection{Network Science}\label{sec:netsci}
Network science has gained a lot of interest for many biological and social science applications. However, to the best of our knowledge, network concepts have \textit{not} been used to understand the inner workings of deep neural networks. To that end, several ideas from network science can be used for better understanding deep network architectures.

\noindent
\textbf{Betweenness Centrality:}
Given a network $\mathcal{G}=\{\mathcal{V}, \mathcal{E}\}$, betweenness centrality $B(v)$ of a node $v \in \mathcal{V}$ is a measure of how central a node is in the network. Specifically, $B(v)$ computes how many shortest paths between different pairs of nodes in the network pass through $v$. Mathematically, $B(v)$ is expressed as:
\begin{equation}
  \vspace{-3px}
    B(v) = \sum_{s\neq t \neq v} \frac{\sigma_{st}(v)}{\sigma_{st}}
    \vspace{-3px}
\end{equation}
where, $\sigma_{st}$ is the number of shortest paths between nodes $s, t \in \mathcal{V}$, and $\sigma_{st}(v)$ are the shortest paths through $v$ ~\cite{newmanComm}.

\noindent
\textbf{Community Structure:}
Communities in a network refer to groups of tightly connected nodes. Intuitively, a community can be defined as a group of nodes where the number of connections within this group is significantly higher than what we would expect in a randomly connected network. Mathematically, communities can be computed by maximizing a modularity function~\cite{newmanComm}:
\begin{equation}
    \max_{\mathbf{g}=\{g_1, g_2, \ldots, g_k\}}\ \  \frac{1}{2m} \sum_{ij} \Bigg[A_{ij} - \frac{1}{\gamma} \cdot \frac{k_i k_j}{2m}\Bigg]\delta(g_i, g_j)
    \label{comm}
\end{equation}
where, $m$ is the number of edges, $k_i$ is the degree (number of connections) of node $i$, $A_{ij}$ is the weight of the link between nodes $i$ and $j$, and $\delta$ is Kronecker delta. The idea is to find groups of tightly connected nodes, $\mathbf{g}=\{g_1, g_2, \ldots, g_k\}$, that map the nodes $\mathcal{V}$ to $k$ communities. The $k_i k_j/2m$ factor represents the number of links one would expect in a randomly connected network. Finally, $\gamma$ controls the resolution of communities: lower gamma will detect a higher number of smaller communities.

\subsection{Interpretability}
Current interpretability techniques fall into two classes. The first class consists of gradient-based methods that compute the gradient of the output with respect to the input, treating the gradient flow as a saliency map \cite{int}. The other type leverages perturbation-based techniques to approximate a complex model using a locally additive model, thus explaining the difference between test output-input pair and some “reference” output-input pair. For instance, \cite{shap} proposed SHAP, a class of methods that randomly draws points from a kernel centered at the test point and fits a sparse linear model to locally approximate the decision boundary. While gradient-based techniques like \cite{int} consider infinitesimal regions on the decision surface and take the first-order term in the Taylor expansion as the additive model, perturbation-based additive models use the finite difference between the input and a reference vector. 

\subsection{Information Theory}\label{sec:IT}
Mutual information has proven to be a valuable tool for feature selection at training time leveraging dimensionality reduction \cite{fsmi}; it can be represented as: $$I(X; Y) = \mathbb{E}_{(x,y)} \left [- \log\left( \frac{\mathbb{P}(x,y)}{\mathbb{P}(x)\mathbb{P}(y)}\right)\right ]$$

\noindent
where $\mathbb{E}_{(x,y)}$ is the expectation over x and y, and $\mathbb{P}(x,y)$, $\mathbb{P}(x)$, $\mathbb{P}(y)$ represent their joint and marginal distributions.

More recent work has represented deep neural networks as Markovian chains to create an information bottleneck theory for deep learning \cite{opentishby}. However, these works do not tackle the interpretability problem directly.
Previous works look to find $I(X; Y)$, the mutual information between an input vector and the output vector. In order to explain the conditional distribution of the output vector given the input vector, some have developed an efficient variational approximation to mutual information \cite{l2x}. However, this model fails to recover the per-feature mutual information, a requisite of our model to explain how information flows through all possible paths. As such, we leverage \textit{point-wise mutual information} (PMI) estimates that would assign a real-valued quantity to each edge for an individual sample; essentially, we remove the expectation from a normal mutual information formulation and are left with the negative log of the joint over the marginals of the observed values. Using PMI also helps us recover realistic feature attributions, as shown qualitatively hereafter.
To estimate the mutual information between two empirical continuous distributions (i.e., the activations at nodes), we use the mixed continuous-discrete KSG estimator~\cite{gao2017estimating}, the EDGE estimator~\cite{noshad2019scalable}, and the MINE estimator~\cite{mine}.

\section{Proposed Approach}
\begin{figure}[h!]
    \centering
    \includegraphics[width=4cm]{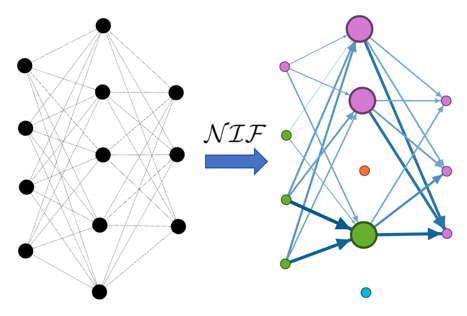}

    \caption{Traditional Model to NIF Network. Color of nodes corresponds to communities. Size of nodes corresponds to betweenness centrality.}
    \label{fig:NIF}
\end{figure}

Our proposed NIF approach transforms a traditional deep learning model into a representation that actually captures the information-theoretic relationship between nodes (Fig \ref{fig:NIF}). 
Our approach extends the approach in \cite{mine} and decomposes their approximation of $I(\mathcal{X};\mathcal{Z})$ to give us $I(\mathcal{X}_i;\mathcal{Q}_k)$, where $X_i$ is a dimension of $X$ (specifically the $i^{th}$ feature of the input vector), and $Q_k$ is any quantity of interest (perhaps the $k^{th}$ neuron in a hidden layer or a class of the output vector).  
Assuming that mutual information is composable and entropy is non-decreasing, we can calculate the mutual information for any feature $\mathcal{Q}_k$ by leveraging a tractable approximation from \cite{fsmi}:
\begin{equation}
    \vspace{-7px}
    \mathcal{NIF} = I(\mathcal{X}_i;\mathcal{Q}_k) = I(\mathcal{X};\mathcal{Q}_k) - \beta\sum_{j=1}^{i-1} I(\mathcal{X}_i;\mathcal{X}_j)
    \label{aa}
    \vspace{-2px}
\end{equation}
where parameter $\beta$ can be used to tune the removal of mutual information between features. The first term is the \textit{relevance} of $\mathcal{X}$ to $\mathcal{Q}_k$ and the second term is \textit{redundancy} in interactions between dimensions of the input. $\beta$ is chosen via cross-validation and was set at $5\times10^{-4}$ for our experiments. Note that the complexity of our estimator scales linearly by network depth, but quadratically by layer dimensionality.

\subsection{Feature Attribution}
NIF recovers a feature attribution, as outlined in Algorithm~\ref{alg:example}. We find all the possible paths between a feature of interest $x_i$ and any of the outputs $y_1, \ldots, y_c$. To find out the value of a path, we take the product of all NIF calculations along the path. We then sum over all of the possible values to find $\mathcal{A}_{i,j}$, our desired feature attribution for feature $x_i$ and class $y_j$. Mathematically, the element $\mathcal{A}_{i,j}$ of our \textit{attribution matrix} $\mathcal{A} \in \mathbb{R}^{n\times c}$ (where $n$ is the number of features and $c$ is the number of classes) is:
\begin{equation}
\vspace{-3px}
    \mathcal{A}_{ij} = \sum_{p\in \mathbb{P}}\ \ \ \prod_{l \in \mathbb{L}}\mathcal{NIF}_p(l)
\label{eq:NIF}
\vspace{-3px}
\end{equation}
where $\mathbb{P}$ is the set of all directed paths from input $x_i$ to class $y_j$ in the neural information flow network, and $\mathbb{L}$ is the set of links on each path $p\in \mathbb{P}$.

\begin{algorithm}[h!]
   \caption{NIF Attribution for predictor $f$}
   \label{alg:example}
\begin{algorithmic}
   \STATE {\bfseries Input:} NIF model for learnt predictor $\mathcal{NIF}_{\widehat f}$, feature index $i$, class of interest $j$
   \vspace{4pt}
   \STATE Find the set of links $\mathbb{L}$ (where each link contains two nodes) for all possible paths $\mathbb{P}$ between feature $i$ and class $j$ (that is $x_i$, $y_j \in {p} \subset \mathbb{P}$)
   \vspace{4pt}
   \FOR{path $p \in \mathbb{P}$}
   \FOR{link $l \in \mathbb{L}_p$}
   \STATE Compute the NIF between the two links $\mathcal{NIF}_p(l)$
   \vspace{4pt}
   \ENDFOR
   \vspace{4pt}
   \ENDFOR
   \vspace{4pt}
   \STATE {\bfseries Output:} Find the sum of $\mathcal{NIF}_p(l)$ across all paths $\mathbb{P}$ multiplied over all links on the path $\mathbb{L}_p$
\end{algorithmic}
\end{algorithm}
\vspace{-20px}
\subsection{Extension to Convolution Neural Networks}
Moving beyond feed forward networks, we have only addressed global explanations (e.g., which neuron affects which other neuron) because we estimate mutual information that, by definition, averages across all samples. However, for feature attribution in image classification problems via CNNs, we need instance-wise explanations such as saliency maps \cite{baehrens2010explain}. Therefore, we formulate a point-wise mutual information (PMI)-based technique to generate saliency map-based explanation for individual samples. 
Recall that PMI can be obtained by removing the mean over all samples (see Section~\ref{sec:IT}) to preserve sample-specific information. Since the activations at channels of a CNN are continuous, we need a continuous estimate of the PMI. Towards this end, KSG estimator~\cite{gao2017estimating} estimates mutual information as a mean (over samples) of the log of the Radon-Nikodym derivative: 

\begin{equation}
    \hat{I}(X; Q) = \frac{1}{N}\sum_{i=1}^N- \log\left( \frac{d\mathbb{P}(x_i,q_i)}{d\mathbb{P}(x_i)\mathbb{P}(q_i)}\right)
\end{equation}

\noindent
Hence, PMI can be directly obtained by the negative log of the Radon-Nikodym derivative above. Hence, from~\cite{gao2017estimating}, we can estimate PMI between two quantities of interest $Q_j$ and $Q_k$ (e.g., between channels $Q_j$ and $Q_k$) for sample $i$ is: 
\begin{equation}
    \mathcal{NIF}_{Q_j, Q_k} = \psi(k) + \log(N) - \alpha
    \label{try}
\end{equation}
where $\alpha = \log(n_{q_j}+1) + \log(n_{q_k}+1)$, $\psi(\cdot)$ is a digamma function, $k$ is a parameter to specify the number of nearest neighbors, $N$ is the total number of samples, $n_{q_j}$ ($n_{q_j, i}$) is the number of samples of $Q_j$ within some distance $\rho$ (see~\cite{gao2017estimating}). Next, we describe how we obtain saliency maps for CNNs.\vspace{1mm}

\noindent
\textbf{Saliency Maps for CNNs:} The saliency maps show which part of the input image are important for the final prediction. Hence, the NIF links for CNNs must map the input image to the channels. Therefore, because the first convolutional layer maps images to multiple channels, we compute NIF from individual pixels to output channels for the first layer. This stage incurs most of the computation. Note that, in CNNs, the interpretation is w.r.t. output \textit{channels} (and not individual neurons -- that would be merely a pixel of an output channel\footnote{In the context of image classification problems, we always talk about what a channel represents in the final convolutional layer of a CNN, e.g., for a CNN that classifies between cats and dogs, some channels activate for dog-faces, others activate for cat-stripes, etc.~\cite{dd}}). Therefore, to get $Q_j$, $Q_k$ in the rest of the network, we use averaged outputs of channels in each convolutional layer. Consequently, creating NIF links for a CNN becomes analogous to that in a feedforward neural network (since each channel now represents a single averaged output), except for the very first layer, that maps input pixels to the first channels. 
Finally, Equation~\eqref{eq:NIF} is used to traverse the NIF links throughout the CNN and to obtain a saliency map. Specifically, the links map the output to the input pixels and the resulting attribution matrix is used as a saliency map.

\vspace{-8px}
\section{Results}
\vspace{-8px}
In order to test the fidelity of NIF, we run a few experiments that validate our proposed technique. We run all experiments on UCI datasets, namely Iris and Banknote authentication, both of which provide us with a small enough feature space to interpret and visualize the network \cite{uci}.
\vspace{-10px}
\subsection{Network Visualization} We start by visualizing NIF for a one layer perceptron trained on the Banknote dataset with ReLU activations and optimized via ADAM \cite{kingma2014method}, and a two layered network with ReLU activation on the Iris dataset.
\begin{figure}[h!]
\centering
\subfigure[][]{%
\label{fig:ex3a}%
\includegraphics[width=3.1cm]{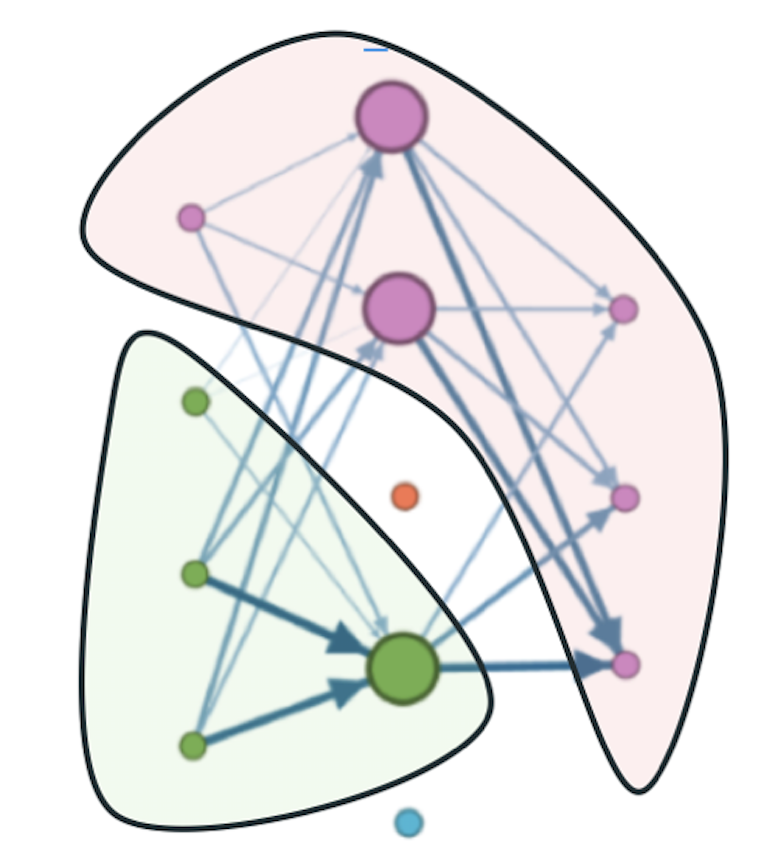}}%
\hspace{8pt}%
\subfigure[][]{%
\label{fig:ex3b}%
\includegraphics[width=3.1cm]{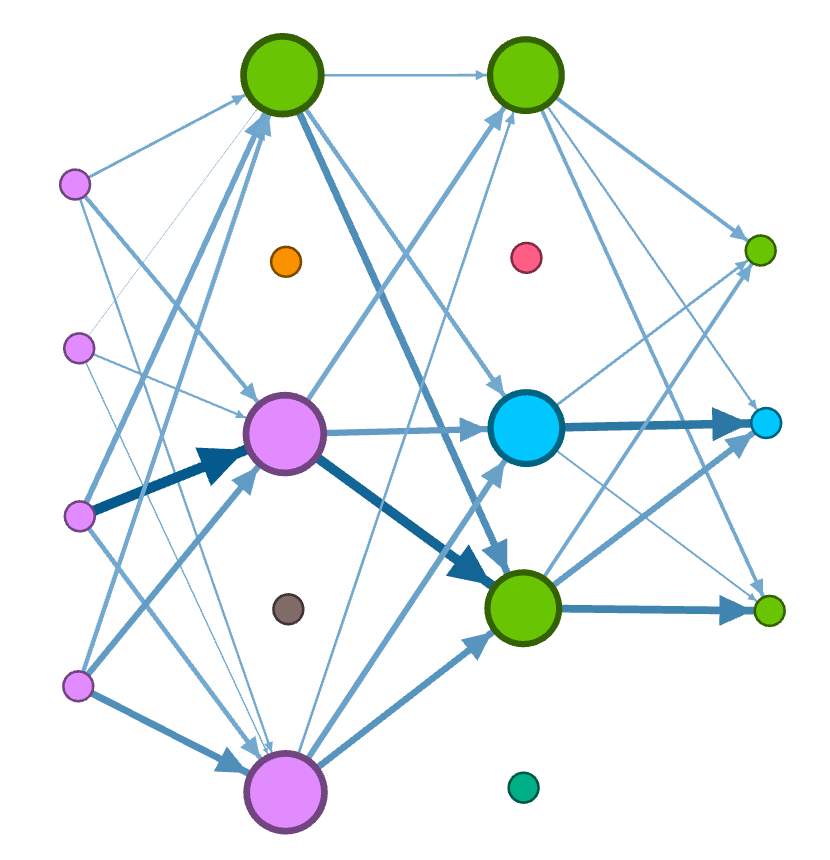}}
\vspace{-10px}
\caption[]{Visualizing NIF models trained  with ReLU activation and an ADAM optimizer \cite{kingma2014method} \subref{fig:ex3a} NIF network for an Iris perceptron with the communities highlighted
\subref{fig:ex3b} NIF network for a multi-layer perceptron trained on the Iris dataset}
\label{fig:i}%
\vspace{-10px}
\end{figure}
In Figure~\ref{fig:i}, we show the NIF network created using Equation~\eqref{try}. For both NIF models, we normalize the information flow per layer to ease visualization of the edges. The thickness of an edge denotes how much information is flowing between any two nodes: the thicker the connection, the more information travelling form one node to the next. The size of the node denotes its centrality: the bigger the node, the more central it is for information to propagate through the network. The color of the node denotes which community the node is a member of: for standard resolution of $\gamma = 1$, we use Equation~\eqref{comm} to find the communities outlined in Figure~\ref{fig:ex3a}. In Figure~\ref{fig:ex3b}, of the five hidden neurons in the final hidden layer, only three are central to the model's final prediction. This makes sense as the ReLU activation at those nodes is zero: thus, ReLU effectively stifles information from flowing through the network. 

\vspace{-10px}
\subsection{Network Pruning}
It is worthwhile to note that both of the models described above received upwards of 96\% accuracy on a held-out test set. 
Though the state of the art network pruning work has been established \cite{louizos2018learning,Theis2018a}, we show how pruning arises from a NIF model and compares to the norm in the pruning literature. 
We ran experiments wherein we zero out the weights and biases of the original model for zero activation neurons in the original model. 
\begin{figure}[h!]
\vspace{-7px}
\centering
\subfigure[][]{%
\label{fig:ex3-a}%
\includegraphics[width=3.8cm]{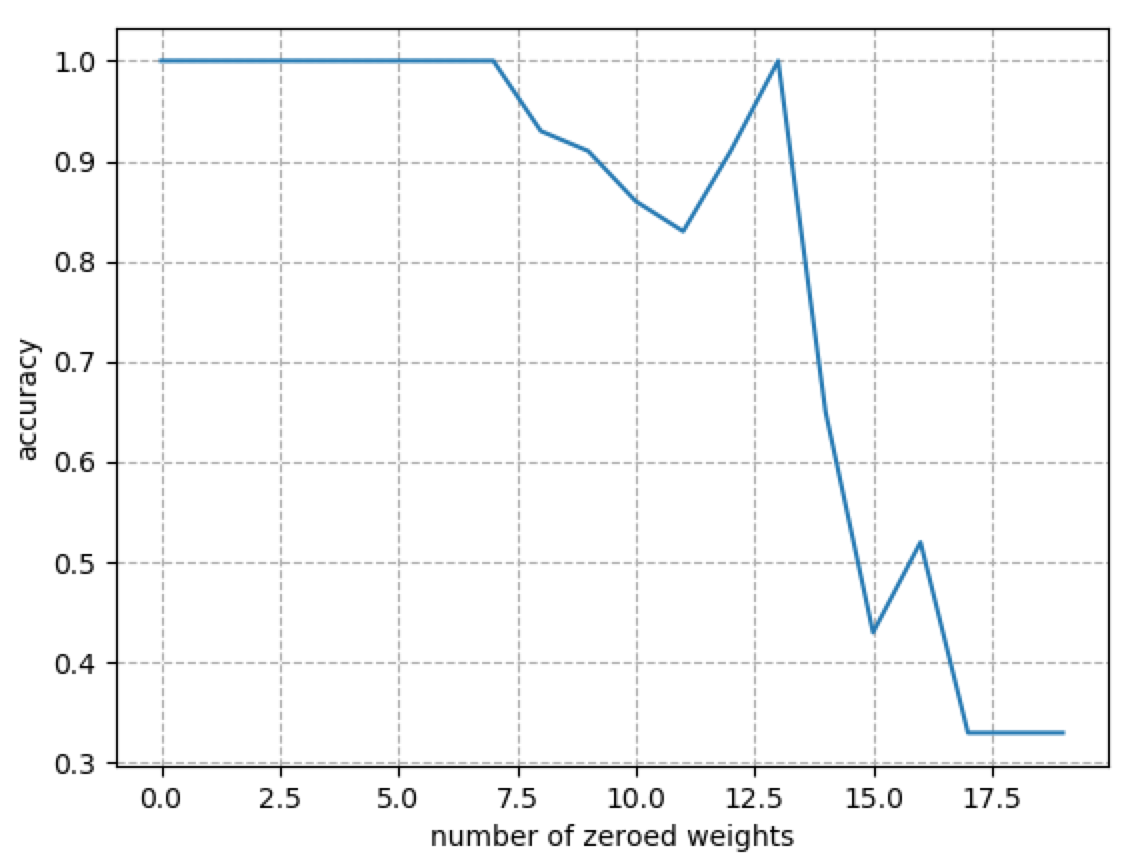}}%
\hspace{8pt}%
\subfigure[][]{%
\label{fig:ex3-b}%
\includegraphics[width=3.8cm]{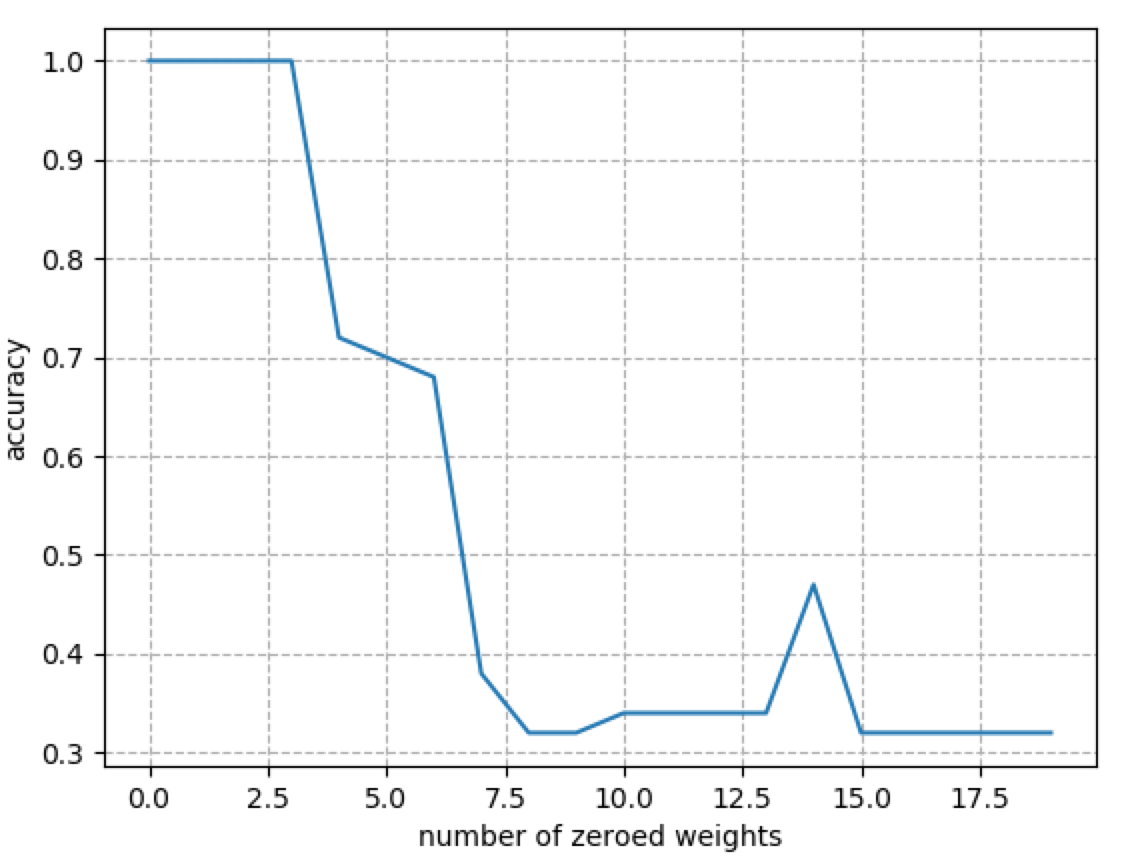}}
\vspace{-7px}
\caption[]{Accuracy vs Number of Zeroed Weights plot for each of the NIF models in Figure~\ref{fig:i} respectively}
\label{fig:ii}%
\vspace{-5px}
\end{figure}
In Figure~\ref{fig:ii}, we report accuracy as the weight matrices of the original network are made more sparse. As the we zero out the links with low NIF importance, we do not lose much accuracy; however, the decrease in accuracy levels out around the 33\%, as expected from three class problem.

\subsection{Feature Attribution}
\begin{figure}[h!]
\centering
\subfigure[][]{%
\label{fig:ex4-a}%
\vspace{-10px}
\includegraphics[width=3cm]{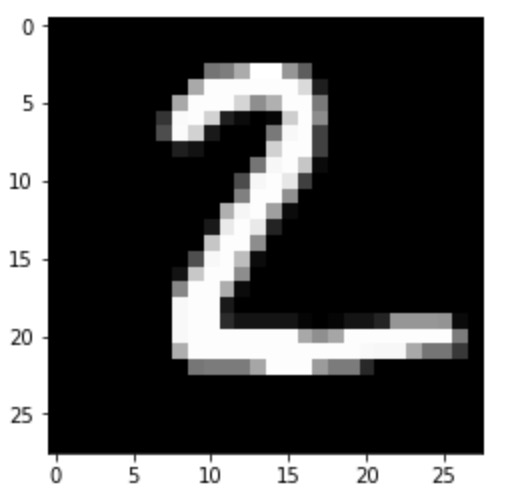}}%
\hspace{8pt}%
\subfigure[][]{%
\label{fig:ex4-b}%
\includegraphics[width=3cm]{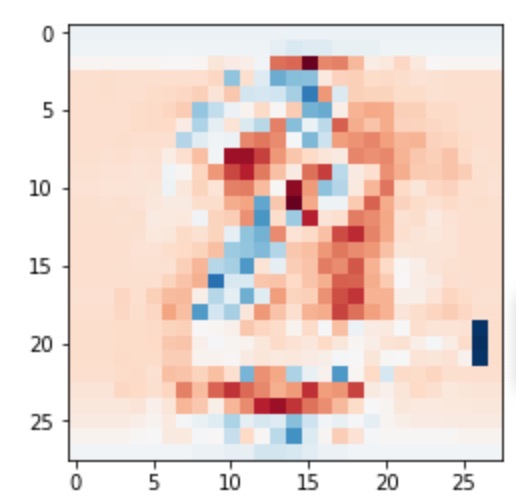}}
\vspace{-10px}
\caption[]{NIF attribution: Figure~\ref{fig:ex4-a} contains the original input and Figure~\ref{fig:ex4-b} contains a feature attribution found using Algorithm~\ref{alg:example} (red means negative and blue means positive)}
\label{fig:iv}%
\vspace{-10px}
\end{figure}

To obtain a NIF feature attribution for a CNN trained on MNIST using Equation~\eqref{try}, we compare NIF against a few popular feature attribution techniques: SHAP \cite{shap} and Integrated Gradients \cite{int} in Table~\ref{attribution}. Using the two sample Kolmogorov-Smirnov test for goodness of fit between two empirical distributions (here, the raw mutual information attribution between the input and output classes and the attribution in question), we find that NIF outperforms these existing techniques, which means NIF is likely drawn from the same distribution as the raw mutual information. In Figure~\ref{fig:iv}, we validate NIF's performance by providing a qualitative example of a NIF attribution, obtained via Equation~\eqref{try}. Thus, an information theoretic feature attribution is viable. 
\begin{table}[h!]
\label{tab}
\begin{center}
\begin{small}
\begin{sc}
\begin{tabular}{lccc}
\toprule
Attribution &  K-S statistic & p-value \\
\midrule
NIF &  1.0 & 0.011\\
SHAP &  0.75 & 0.107\\
Integrated Gradients & 0.25 & 0.996\\
\bottomrule
\end{tabular}
\end{sc}
\end{small}
\end{center}
\vspace{-10px}
\caption{Feature attribution comparison}
\label{attribution}
\end{table}
\vspace{-10px}

\vspace{-5px}
\section{Conclusion}
\vspace{-5px}
We proposed NIF, Neural Information Flow, a new technique for measuring information flow through deep learning models. Merging mutual information estimation with network science, we find that NIF not only provides insight into which pathways are crucial within a network, but also allows us to leverage fewer parameters at inference time, since we can remove parameters deemed useless by the NIF without loss of accuracy. Finally, NIF recovers an information theoretic feature attribution that aligns with existing techniques.

For future work, NIF provides a unique approach to generic function learning. Suppose we want to learn $f^*: \mathbb{R}^d \rightarrow \{0,1\}$ from data $X$ and labels $Y$. Unfortunately, we cannot tell if a trained predictor, $\widehat f$, learned the desired, $f^*$. NIF can clear that doubt by providing insight into the properties of the function learned; particularly, NIF can prune the model class of $f$. Providing theoretical analysis of how certain network properties (lack of centrality or low clustering coefficient) can give us a more restricted model class for $f$ by decreasing the VC-dimension of the hypothesis space over which we trained the predictor $\widehat f$. As such, NIF can help assess which models provide sounder approximations of $f^*$.

\label{sec:refs}

\bibliographystyle{IEEEbib}
\bibliography{strings}

\end{document}